\documentclass[final]{cvpr}

\usepackage{booktabs}
\usepackage{times}
\usepackage{epsfig}
\usepackage{graphicx}
\usepackage{amsmath}
\usepackage{amssymb}
\usepackage{algorithm}
\usepackage[noend]{algpseudocode}
\usepackage{setspace}

\usepackage[pagebackref=true,breaklinks=true,colorlinks,bookmarks=false]{hyperref}

\def\cvprPaperID{****} 
\def\confYear{AI604 2021 Fall}

\def\algname{EoREN}

\begin{document}

\title{Edge-oriented Implicit Neural Representation with Channel Tuning }

\author{Wonjoon Chang\thanks{Contributed equally}, Dahee Kwon\footnotemark[1], Bumjin Park\footnotemark[1]}



\maketitle

\begin{abstract}
Implicit neural representation, which expresses an image as a continuous function rather than a discrete grid form, is widely used for image processing. Despite its outperforming results, there are still remaining limitations on restoring clear shapes of a given signal such as the edges of an image. In this paper, we propose Gradient Magnitude Adjustment algorithm which calculates the gradient of an image for training the implicit representation. In addition, we propose Edge-oriented Representation Network (EoREN) that can reconstruct the image with clear edges by fitting gradient information (Edge-oriented module). Furthermore, we add Channel-tuning module to adjust the distribution of given signals so that it solves a chronic problem of fitting gradients. By separating backpropagation paths of the two modules, EoREN can learn true color of the image without hindering the role for gradients. We qualitatively show that our model can reconstruct complex signals and demonstrate general reconstruction ability of our model with quantitative results. 
\end{abstract}
\let\thefootnote\relax\footnotetext{Preprint}

\section{Introduction}
Implicit Neural Representation learns a function, neural network in most cases, which takes pixel coordinates as input and output the corresponding RGB pixel values. 
The implicit problem formulation takes form:
\begin{equation}
    F(\mathbf{x}, \Phi, \nabla_{\mathbf{x}} \Phi, \nabla^2_\mathbf{x} \Phi, \cdots) = 0 , ~~ \Phi : \mathbf{x} \mapsto \Phi(\mathbf{x})
\end{equation}
where $\Phi$ is the function we are interested and $F$ is the relation between $\mathbf{x}$ and $\Phi$.  For example, a single image data can be represented as  $D=\{ (\mathbf{x}_i, f(\mathbf{x}_i) \}$ where $\mathbf{x}_i$ is pixel coordinates associated with their RGB colors $f(\mathbf{x}_i)$. $\Phi(\mathbf{x})$ can be enforced to output RGB colors and minimize  $||f(\mathbf{x}) - \Phi(\mathbf{x})||^2$. The goal of the implicit problem is to find the function $\Phi$.

Representing image as a continuous function can have benefits in terms of efficiency and flexibility which makes Implicit Neural Representation be widely used in image reconstruction and super resolution task. For example, Sitzmann et al. \cite{NEURIPS2020_53c04118} achieved breakthrough performance on image reconstruction by using implicit representation with sine activation. However, it does not imply implicit neural representation methods can reconstruct images perfectly. 

\begin{figure*}[ht]
\centering
\includegraphics[width=16cm, trim=0cm 0cm 0cm 0cm, clip]{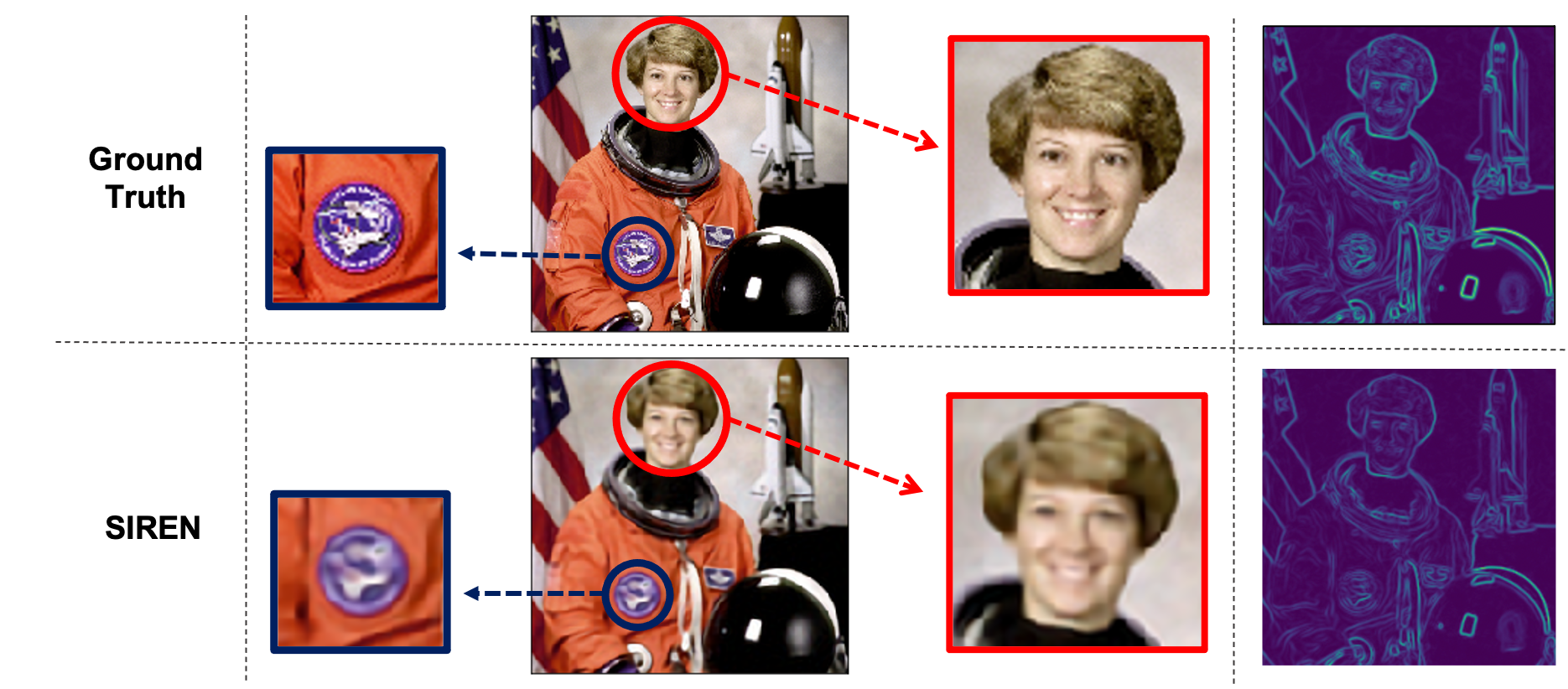}
\caption{First row (top) is an original image and the corresponding gradient values. Second row shows the result from SIREN model trained with pixel values. We enlarge some points to clearly show the detailed differences.}
\label{SIREN comparison}
\end{figure*}

In Figure~\ref{SIREN comparison}, we show some drawbacks of SIREN, proposed on \cite{NEURIPS2020_53c04118}. On the second row of Figure~\ref{SIREN comparison}, the overall reconstruction seems quite plausible, but you can see that local features such as eyes and the logo on the jacket are not well caught. This phenomenon is more pronounced in the gradient visualization. In other words, RGB target based learning is lack of edge reconstruction while edge target based learning is lack of color reconstruction. Yet, the above two are not trade-off properties but tasks that must be considered simultaneously. \\
\indent Thus, in this paper, we propose edge-oriented implicit representation with color-tuning architecture (EoREN) not only detect the edge clearly but also reproduce the color well. EoREN is supervised in a direction that gradients trained with implicit function become close to true gradients, which prioritizes fitting gradients rather than pixel values.
In the process, it is required to approximate the true gradient values because we couldn't directly access to the ground truth gradients. Generally, directional filters such as sobel filter are widely used which amplify the total magnitude of the gradients. 
Since the enlargement of gradients makes the model be trained with wrong target values, we propose gradient magnitude adjustment (GMA) that allows the model to learn in a more accurate direction. In addition, for more accurate color reconstruction, channel tuning module that adjusts the scale of the pixel values are added after gradient fitting.


\section{Related Work}
\textbf{Implicit neural representation.}  Implicit neural representation is a task of learning a function which puts position of the data as inputs and outputs the signal of the position. Recently, many researchers try to adopt implicit neural representation into various deap learning fields \cite{genova2020local, Michalkiewicz_2019_ICCV} such as 3D reconstruction \cite{10.5555/3454287.3454388, gropp2020implicit, Atzmon_2020_CVPR}. 
The Multi Layer Perceptron (MLP) is mostly used to train the implicit function with ReLU activation. However, Sitzmann et al. \cite{NEURIPS2020_53c04118} pays attention to the fact that MLP with ReLU activation is ineffective to represent the signal data and suggests SIREN model which uses periodic activation function that can learn the representation with image, audio, and video signal well. Also, instead of training separate neural network for each image, more generalized way is proposed with feature representation of images \cite{chen2019learning}. Chen et al. \cite{chen2021learning} suggests Local Implicit Image Representation and shows a way to train continuous representation learning method in super resolution task. However, none of these approaches examines the effect of the complexity of images when training implicit representation function. Different from these works, our work focuses on the capability of the implicit neural representation. 

\textbf{Gradient-based edge detection.} Edge detection is an important step that becomes the cornerstone of image processing for object detection and recognition. Traditionally, gradient information was mainly used for edge detection \cite{shrivakshan2012comparison, katiyar2014comparative}. The gradient of the image refers to amount of changes from a low brightness value to a high value or vice versa. Therefore, finding pixels with large gradients can be seen as finding the boundaries of the objects in the image. The most commonly used gradient-based edge detection method is  \cite{canny1986computational} called Canny edge detection, which uses Gaussian filtering and the Sobel mask filter. Canny detection method is simple and effective, yet has the disadvantages of time-consuming and parameter adjusting issue. In addition, processing complex images with only edge lacked information. 

Along with these limitations, traditional methods have been overshadowed by the popular deep learning methods such as CNN for a while. \cite{xie2015holistically} utilizes the feature maps of intermediate layer for edge detection and \cite{ledig2017photo} uses generative adversarial network for super-resolution task. Recently, the Neural Implicit Representation based models \cite{mildenhall2020nerf, NEURIPS2020_53c04118} boost the performance of image reconstruction and super-resolution task by considering continuous representation space. Even though these deep learning based methods could process images very well with their super-complex structure, there still exist some challenges to be solved in the point of the exact image reproduction. One of them is to recognize the edge of objects and restore them clearly. So in order to overcome the performance bottleneck, many studies applied the gradient-based edge detection method to deep neural network. \cite{ma2020structure} uses the gradient information to improve the performance of the existing super-resolution method \cite{ledig2017photo}. And in addition, various techniques such as \cite{shibata2016gradient} uses gradient-based edge detection method to improve the performance of image restoration. We also focus on improving the reconstruction performance of the existing neural implicit presentation by utilizing the gradient information.

\section{Background}
\subsection{Signal Fitting with Implicit Representations}

Implicit neural representations learn implicitly defined functions to model many different types of signals such as image, video, and audio processing in a continuous, memory-efficient way. The following equation is a general form of loss to train the implicit function $\Phi$ to represent the signal $f$ on $\Omega$:
\begin{equation}
    \mathcal{L}=\int_{\Omega}
    \Vert \Phi(\mathbf{x})
    - f(\mathbf{x})
    \Vert d\mathbf{x}.
    \label{eq-inr}
\end{equation}

It turns out that sinusoidal representation networks (SIRENs) can effectively learn representing complex natural signals by using periodic activation functions such as sin activation functions~\cite{NEURIPS2020_53c04118}. SIRENs converge significantly faster than existing networks that use other activation functions as well as feature high image fidelity. It shows that reconstructed images from SIRENs have more realistic image properties so that their gradients and laplacians are close to the ground truth.

\subsection{Poisson Equation}

A supervised neural network is generally trained to provide an output that is close to the ground truth. Implicit neural representations also focus on learning implicit functions of given signals. However, reconstructed images from SIRENs that are trained to fit given images may not reflect complex patterns such as edges in a logo. It is because there is no supervision to learn exact edges in images and the network is optimized to fit average color-level of pixels.

The implicit neural representation can also be supervised solely by derivatives of a given signal~\cite{NEURIPS2020_53c04118}. That is, the model is never provided with the actual function values. The Poisson equation is the simplest elliptic partial differential equation (PDE) that requires to fit the first or higher-order derivatives of a given signal. In this problem, an unknown ground truth signal $f$ is estimated from discrete samples of either is gradients $\nabla f$ or Laplacian $\Delta f=\nabla \cdot \nabla f$ as

\begin{equation}
    \mathcal{L}_{\textrm{grad.}}=\int_{\Omega}
    \Vert \nabla_\mathbf{x}\Phi(\mathbf{x})
    - \nabla_\mathbf{x} f(\mathbf{x})
    \Vert d\mathbf{x},
    \label{eq-poisson-grad}
\end{equation}

\begin{equation}
    \mathcal{L}_{\textrm{lapl.}}=\int_{\Omega}
    \Vert \Delta_\mathbf{x}\Phi(\mathbf{x})
    - \Delta_\mathbf{x} f(\mathbf{x})
    \Vert d\mathbf{x}.
\end{equation}

Solving the Poisson equation enables the reconstruction of images from their derivatives so that the implicit function represents edges of images much better. Actually, a way to fit gradients with SIREN for a RGB image has not been directly described in the original paper. We train SIRENs to minimize the difference between automatically computed gradients and true gradients calculated by sobel filters for each RGB channel.

Figure~\ref{SIREN comparison} shows one example of the results. SIREN trained to fit gradients can describe complex patterns such as the logo in the jacket and clearly reconstruct eyes of the astronaut. However, there is remaining intensity variations due to the ill-posedness of the problem.


\subsection{Sobel Filter}
In the SIREN paper, the Sobel filter was used to calculate the ground truth gradient. The Sobel filter is used in image processing and computer vision, particularly within edge detection algorithms where it creates an image emphasizing edges. Technically, it is a discrete differentiation operator, computing an approximation of the gradient of the image intensity function. The Sobel filter is based on convolving the image with a 3x3 filter in the horizontal and vertical direction.
\begin{equation}
\mathbf{G}_x=
\begin{bmatrix}
-1 & 0 & 1\\
-2 & 0 & 2\\
-1 & 0 & 1
\end{bmatrix}
,\quad
\mathbf{G}_y=
\begin{bmatrix}
-1 & -2 & -1\\
0 & 0 & 0\\
1 & 2 & 1
\end{bmatrix} 
\end{equation}

$\mathbf{G}_x$ denotes the Sobel filter for the horizontal direction and $\mathbf{G}_y$ denotes the Sobel filter for the vertical direction. The gradient approximation that the Sobel filter produces is relatively crude, in particular for high-frequency variations in the image. 

\begin{figure}[ht!]
\includegraphics[width=8cm, trim = 0cm -1cm -1cm 0cm, clip]{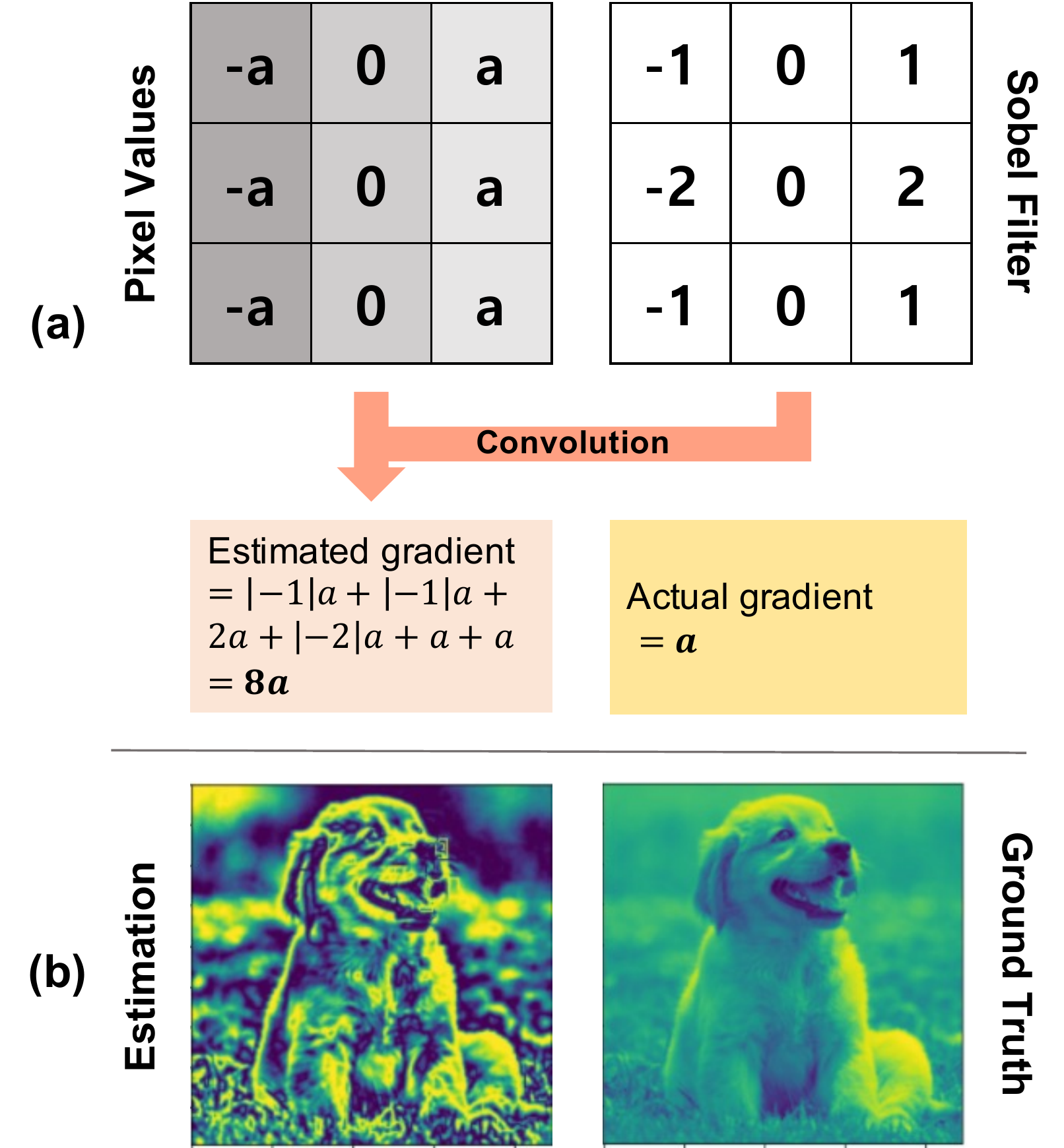}
\centering
\caption{Induced problem of the Sobel filter. (a) shows that the gradients computed by convolving the Sobel filter can be overestimated compared to the original gradient in the simple example. (b) shows that the reconstructed dog image from SIREN trained with these estimated gradients has extremely large differences at the edges of the image.}
\label{fig:sobel-problem}
\end{figure}

\section{Method}

\begin{figure*}[ht]
\includegraphics[width=16cm, trim = 0cm -1cm 0cm 0cm, clip]{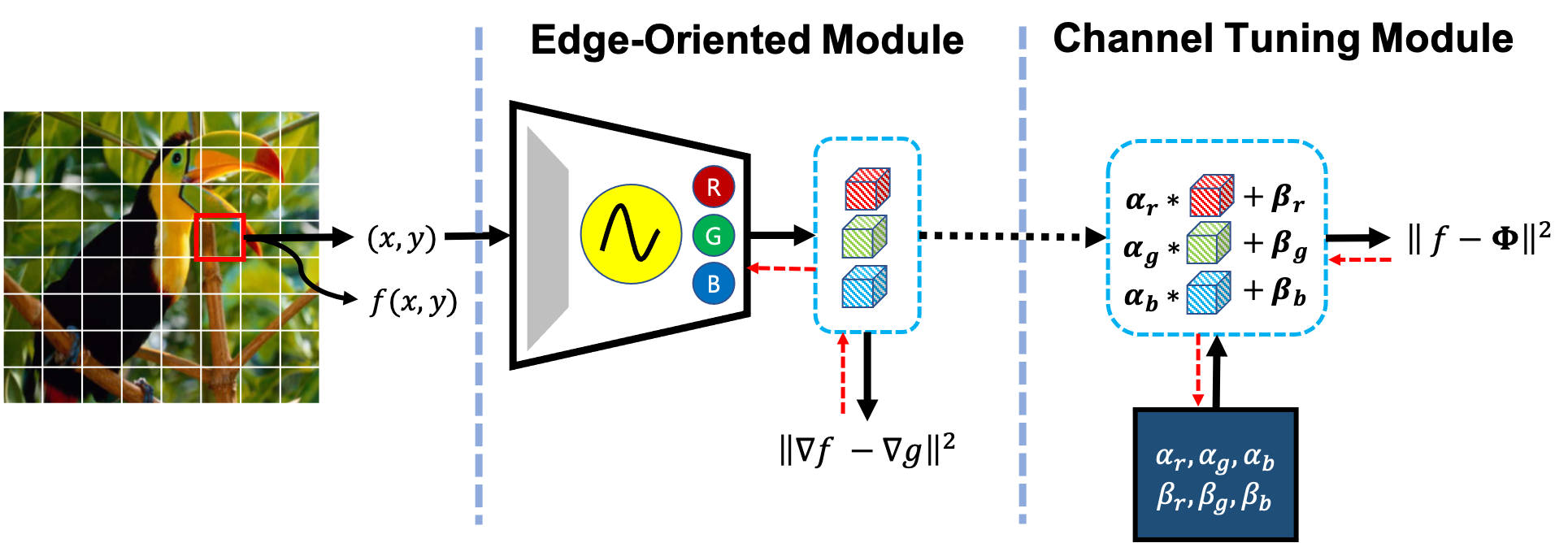}
\centering
\caption{The structure of \algname. Each pixel $(x,y)$ of input image $f$ goes through 1) Edge-oriented module to learn the edge property of the image and 2) channel tuning module to learn the distribution of the pixel values. The loss of Edge-oriented module is computed with the gradient $\nabla f(x,y)$ on pixel $(x,y)$ of the image. The loss of the channel tuning module is computed with RGB value $f(x,y)$. The red dashed line indicates backpropagation.}
\label{fig-eoren-archi}
\end{figure*}

\subsection{Gradient Magnitude Adjustment}

Even though the Sobel filter is widely used as a computationally efficient way to approximate the gradients of images, it is problematic to feed the estimated gradients as the ground truth to the model. Figure~\ref{fig:sobel-problem} describes which problem can be induced from the Sobel filter. Consider the simple case where the gradient value is a constant $a$ in the horizontal direction. When we apply convolution with the Sobel filter, the estimated gradient is $8a$. That is, the Sobel filter overestimate the gradients. To fit the overestimated gradient values, the model sacrifice the correct pixel values such as the dog image in Figure~\ref{fig:sobel-problem} (b). Even a little contrast exists in the original input image, the difference of the values in the edge is overestimated and the reconstructed image becomes very unrealistic.


To address this issue, we propose to adjust the magnitude of gradients computed by filter for realistic image reconstruction. We define the magnitude of the filter as the total sum of the absolute values used in the filter. For instance, the Sobel filter, described in Figure~\ref{fig:sobel-problem}, has magnitude 8 which means that the estimated gradient values are 8 times larger than the actual gradient values. Therefore, we need to offset the gap between the estimated gradients and the actual gradients.

We consider two main components to feed proper gradient values to the model. The first one is the magnitude of the filter which is introduced previously, since it leads to overestimation. The second thing is the size of the image. Let $W^T,H^T$ be the range of the input coordinates and $W,H$ be the width and height of the image respectively. Note that convolving process assumes that the size of the interval between adjacent pixels is 1. However, SIREN uses the different range of input coordinates. As a result, the derivatives of the implicit function with respect to input may have different scale. Therefore, we need to multiply the gradients by $\frac{W}{W_T}, \frac{H}{H_T}$, after normalizing it by filter magnitude. Then, we can feed adjusted gradient values to the model for solving the Poisson Equation.

\begin{algorithm}
\setstretch{1.35}
\caption{Gradient with Magnitude Adjustment}\label{alg:gma}
\begin{algorithmic}[1]
\Procedure{GMA}{$I_{W\times H}$, ${W_T, H_T}$ , $F^x$, $F^y$} 
\State $M_F^x \gets |F^x|$ \Comment{compute filter magnitude}
\State $M_F^y \gets |F^y|$
\State $F^x \gets F^x/M_F^x$ \Comment{normalize by filter magnitude}
\State $F^y \gets F^y/M_F^y$

\State $F^x \gets F^x \times \frac{W}{W_T}$ \Comment{normalize by image size}
\State $F^y \gets F^y \times \frac{H}{H_T}$

\State $grad_x \gets  F^x \circledast I_{W\times H}$ \Comment{convolution}
\State $grad_y \gets  F^y \circledast I_{W\times H}$

\State \textbf{return} $(grad_x, grad_y)$
\EndProcedure
\end{algorithmic}
\end{algorithm}

\begin{figure*}[ht]
\includegraphics[width=16cm, trim = 0cm 0cm 0cm 0cm, clip]{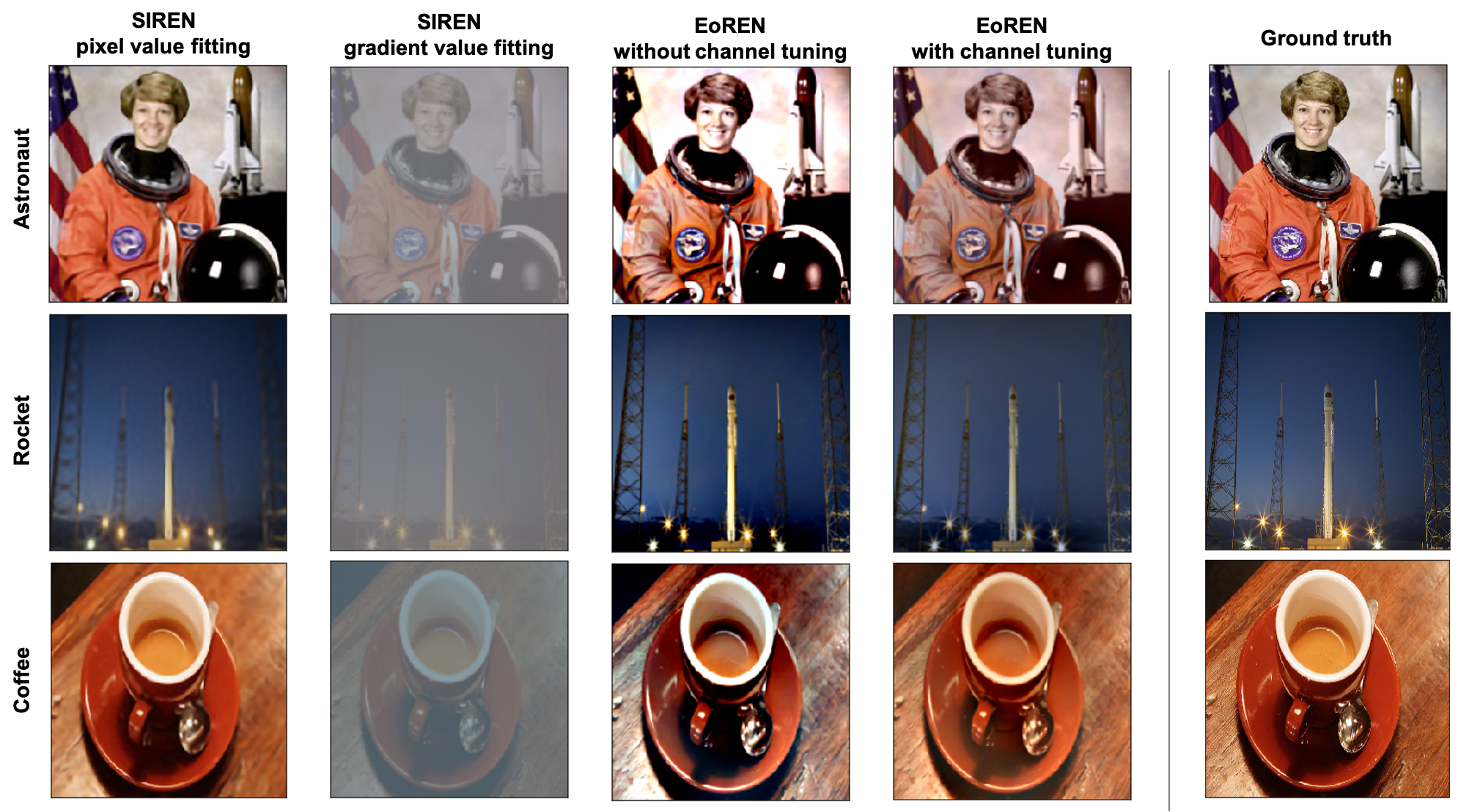}
\centering
\caption{The implicit representation of the RGB images in skimage data with SIREN-pixel, SIREN-grad, \algname~ without the channel tuning module and \algname. }
\label{exp_constructed_rgb}
\end{figure*}

\subsection{\algname}

We add one more additional module to help our model correctly represent pixel values while keeping complex signals by gradient adjustment. We propose Edge-oriented Representation Network  (\algname) that can represent complex patterns in the signal with avoiding the ill-posedness problem through the channel tuning module.
\begin{equation}
\Phi(\mathbf{x}) = (h \circ g) (\mathbf{x})
\end{equation}

\algname~is composed of two functions $g$ and $h$. $g$ represents original implicit representation function of the signal and $h$ is an additional shift transformation for each dimension of outputs. 
\begin{gather}
    g(\mathbf{x}) = \mathbf{W}_n(g_{n-1} \circ g_{n-2} \circ \cdots \circ g_0)(\mathbf{x}) + \mathbf{b}_n \\ 
    g_i: \mathbf{x}_i \mapsto \mathrm{Activation}(\mathbf{W}_i \mathbf{x}_i + \mathbf{b}_i) 
\end{gather}
Here, $g_i : \mathbb{R}^{M_i} \rightarrow \mathbb{R}^{N_i}$ is the $i^{th}$ layer of the network with the weight matrix $\mathbf{W}_i \in \mathbb{R}^{N_i \times M_i}$ and the biases $\mathbf{b}_i \in \mathbb{R}^{N_i}$. The function $g$ is a MLP that acts as an implicit function of the signal. It can be SIREN with the use of sine activation functions.

The objective of our implicit function is not to optimize the values of the signal $f$, but to match the gradients of the signal. Our implicit function $g$ is trained to solve the Poisson equation for fitting true gradients $\nabla_\mathbf{x} f(\mathbf{x})$ for each coordinate $\mathbf{x}\in\Omega$. That is, it minimizes the loss described in Equation~\ref{eq-poisson-grad}.
\begin{equation}
  \mathcal{L}_{\mathrm{grad.}} =  \int_\Omega ||\nabla_\mathbf{x} g(\mathbf{x}) - \nabla_\mathbf{x} f(\mathbf{x})|| d\mathbf{x}
\end{equation}

We call this implicit function \textit{Edge-oriented module}. The Edge-oriented module can be easily supervised to learn complex shapes in a given images. However, since it is optimized only to fit $\nabla_\mathbf{x} f(\mathbf{x})$, $g$ may not follow the actual signal $f$. In the case of image fitting, $g$ may not represent true colors of the image $f$. Therefore, we need to calibrate the distribution of reconstructed images. 

We propose an additional shift transformation layer, called \textit{channel tuning module}, to adjust differences between the output and the target signal. Figure~\ref{fig-eoren-archi} shows the overall architecture of our model. $\mathbf{\alpha}$ and $\mathbf{\beta}$ are vectors of trainable parameters. 
\begin{equation}
h(\mathbf{x}) = \alpha * g(\mathbf{x})+\beta
\end{equation}
It is updated to transform the output of the Edge-oriented module so that it fit the true values of the signal $f$. The loss to optimize $\alpha$ and $\beta$ are as follows:
\begin{equation}
  \tilde{\mathcal{L}} =  \int_\Omega || f(\mathbf{x}) -  \Phi(\mathbf{x})|| d\mathbf{x}
\label{eq-loss-offset}
\end{equation}

Note that we do not backpropagate the loss in Equation~\ref{eq-loss-offset} to optimize $g$. When we set $\alpha=\mathbf{1}$, even if we add the additional channel tuning module, the derivatives of \algname~do not change due to the chain rule, i.e., $\nabla\Phi= \nabla g$. Theoretically, it is enough to set $\alpha=\mathbf{1}$ and only train $\beta$ parameter to fit correct pixel values. However, it has been empirically observed that training $\alpha$ parameter helps the model to fit the distribution of pixel values more precisely. Therefore, in our experiment, we decide to train $\alpha$ in the channel tuning module. Consequently, \algname~can learn true color of the image without hindering the representability for gradients.

\section{Experiment}

In this section, we experimentally show the performance of \algname~model compared to SIREN models with the pixel fitting and the gradient fitting. First, we show the images reconstructed by three models. Also, we conduct quantitative analysis for the Set14 \cite{set14-Huang-CVPR-2015}, DIV2K \cite{div-2k-Timofte_2018_CVPR_Workshops}, and MNIST \cite{mnist-deng2012mnist} dataset with image qualifying metrics, PSNR and SSIM score. Furthermore, we show the result of the image composition task by the \algname~model and demonstrate the advantage and the weakness of \algname. We take an experiment on gray scale of an image because an image with 3 channels requires a bigger size model and more training time. We trained 1000 epochs for each model equally. In the case of \algname~, we trained 950 epochs for the edge-oriented module and 50 epochs for the channel-tuning module.

\subsection{Signal Reconstruction}
We first show the signal reconstruction of our model. We test our model with three color images in skimage data. We choose three images, the astronaut, the rocket and the coffee. The results of reconstruction are described in the Figure \ref{exp_constructed_rgb}. The first two columns in Figure~\ref{exp_constructed_rgb} are the images from SIREN with pixel/gradient fitting. Gradient fitting images successfully represent the edge parts of the image, while they fail to match the RGB values of the image. The results of SIREN with pixel fitting have plausible reconstructed RGB values, but fail to distinguish detailed edges on image, such as the patch in the astronaut image, towers in the rocket image and the background in the coffee image. The images reconstructed by our \algname~are the 3th and 4th columns on Figure~\ref{exp_constructed_rgb}. EoREN without channel tuning module overestimate the edges, such as shadows in the coffee image and dark parts in the rocket image. Then, it results in generating high-contrast image compared to the original images.  Meanwhile, EoREN with channel tuning module shows not only clear but also realistic images with similar color levels compared to the other gradient fitting models.

\begin{figure}[ht]
\includegraphics[width=8cm, trim = 0cm 0cm 0cm 0cm, clip]{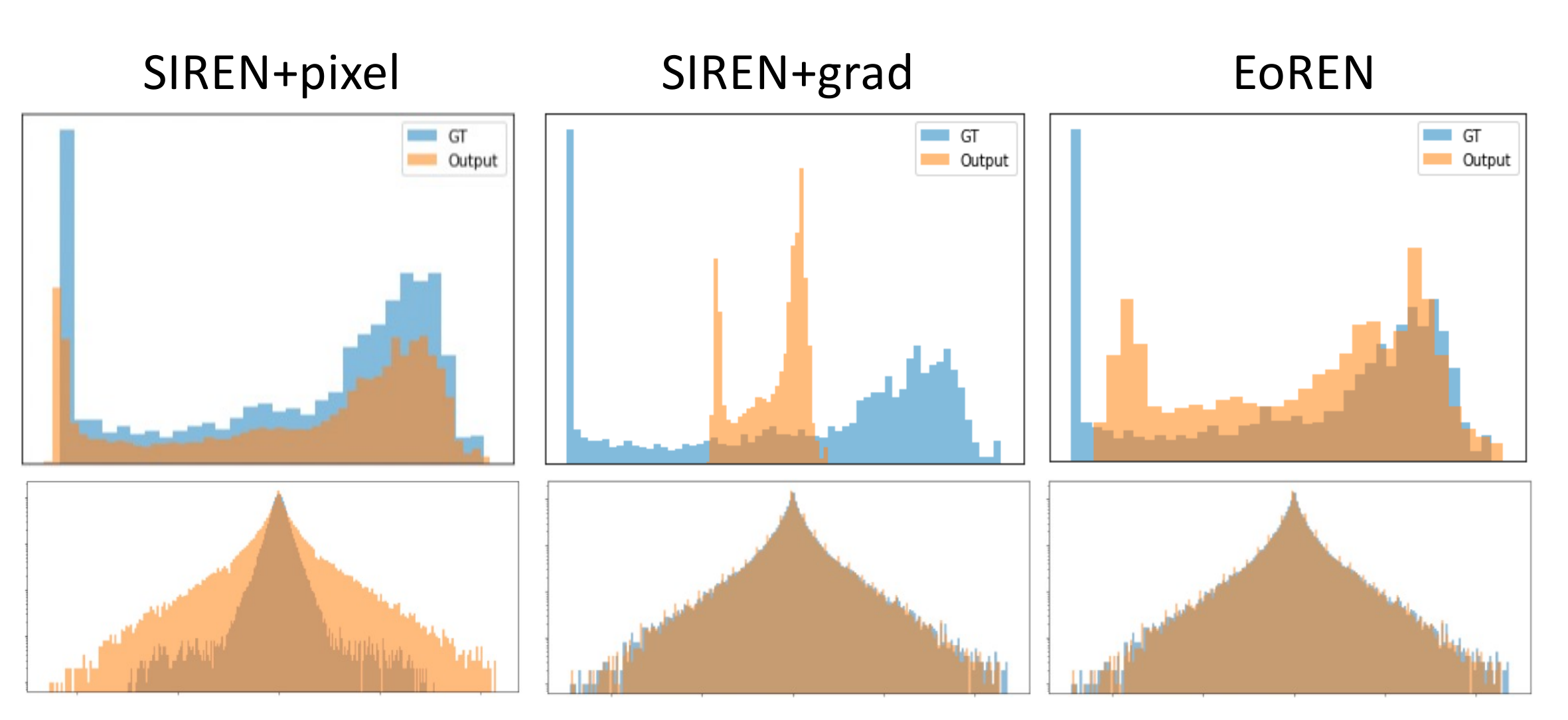}
\centering
\caption{The histogram of the pixel values and the gradients for the RGB astronaut image. }
\label{exp_hist}
\end{figure}

\begin{table*}[ht]
\centering
\caption{The result of image reconstruction for the baseline models and the \algname.}
\begin{tabular}{l  c c c c  c c}
\toprule & \multicolumn{2}{p{4cm}}{Set14} & \multicolumn{2}{p{4cm}}{DIV2K} & \multicolumn{2}{p{4cm}}{MNIST} \\
\cmidrule(lr){2-3} \cmidrule(lr){4-5} \cmidrule(lr){6-7} 
Measure      & PSNR   & SSIM  & PSNR   & SSIM & PSNR   & SSIM \\
\midrule
SIREN+Pixel       & \textbf{73.81} & \textbf{0.78}    & \textbf{74.23} & \textbf{0.83} & 83.69 & 0.91 \\
SIREN+Grad        & 56.46  & 0.33 & 55.45 & 0.31  & 54.54 & 0.74\\
\algname        & 66.01  & 0.46  & 64.36 & 0.41 & \textbf{93.96} & \textbf{0.97}\\
\bottomrule
\label{table_psnr_ssim}
\end{tabular}
\end{table*}

\subsection{Disentangled Property on Edge and Offset}
One advantage of \algname~ is the separation of modules for edge and offset. We show the disentangled property in color and edge construction of our model by the reconstruction of the edge-oriented module and the channel-tuning module separately. 
The distributions of the pixel values and the gradients for three models are shown in the Figure \ref{exp_hist}. 
The SIREN with pixel fitting shows that most of the pixel values are matched. However, the model underestimates the gradients and results in the smooth image as shown in the Figure \ref{exp_constructed_rgb}. The SIREN with gradient fitting matches most of the gradients but fails to reconstruct pixel values. We find that two pixel distributions from the ground truth and the model output are similar in shape. Therefore, our proposed model \algname~ linearly transforms the pixel values by the channel-tuning module and matches the pixel distributions.


\subsection{Benchmark Dataset}
Following the previous work in \cite{NEURIPS2020_53c04118}, we compare \algname~ and SIREN on the benchmark dataset. We evaluate our model based on the PSNR and SSIM metric \cite{5596999} which is widely used metric for image reconstruction. The overall results are in the Table \ref{table_psnr_ssim}.

\subsubsection{Set14}

\begin{figure}[ht]
\includegraphics[width=8cm, trim = 0cm 0cm 0cm 3cm, clip]{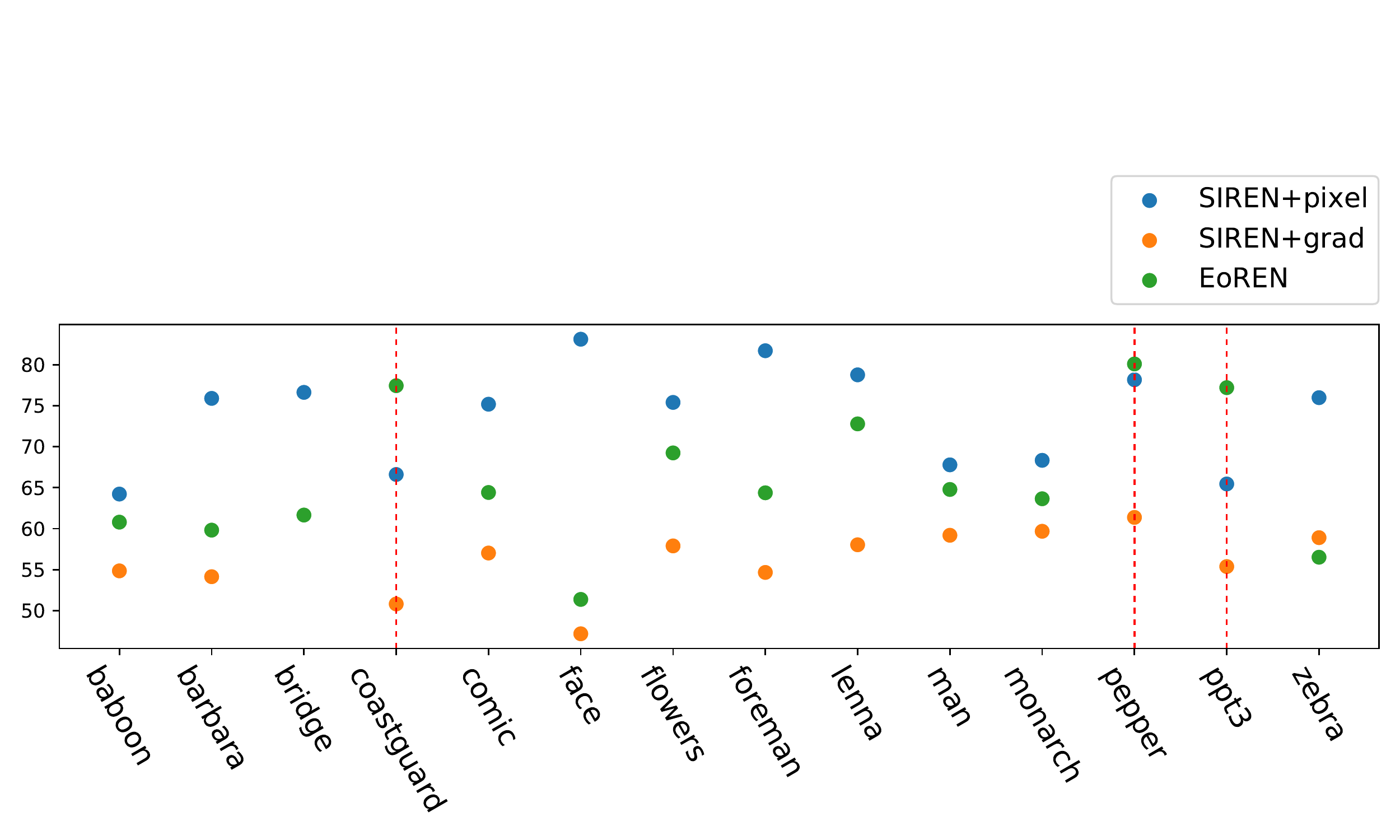}
\centering
\caption{The PSNR performance of SIREN+pixel, SIREN+grad, and \algname~ for Set14 dataset. \algname~ outperforms SIREN+pixel for three images, coastguard, pepper, and ppt3.  }
\label{exp_set14}
\end{figure}

\begin{figure}[ht]
\includegraphics[width=8cm, trim = 0cm 0cm 0cm 0cm, clip]{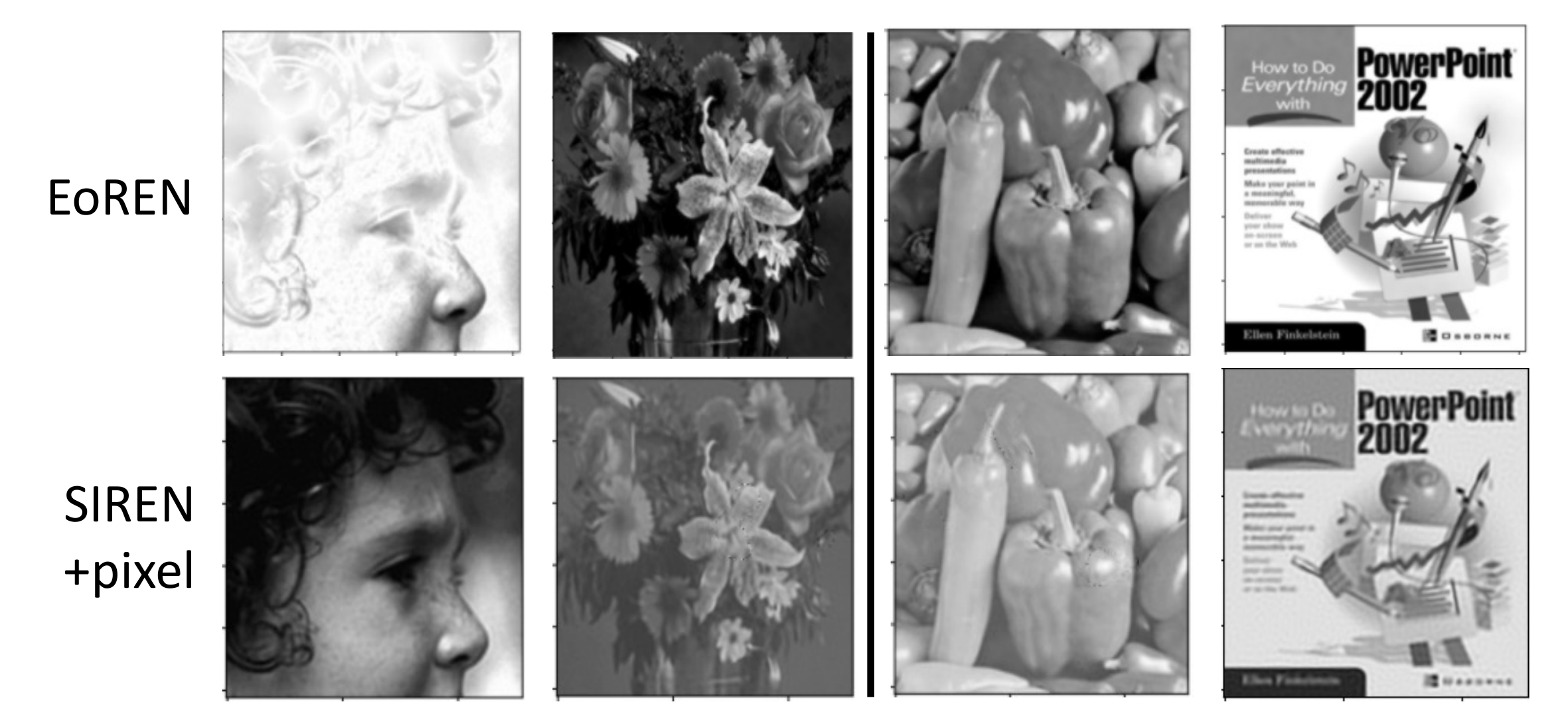}
\centering
\caption{Reconstructed images by \algname~ and SIREN with pixel fitting. the left panel contains the images which SIREN+pixel outperforms \algname, whereas the right panel contains the images which \algname~ outperforms SIREN+pixel.}
\label{exp_set14_eg}
\end{figure}

Set14 contains 14 images from simple to complex. The PSNR performance of \algname, SIREN with pixel fitting, and SIREN with gradient fitting for each sample is in Figure \ref{exp_set14}. We observe that SIREN+pixel shows generally better PSNR scores than \algname. However, there are some samples  on which \algname shows better PSNR than SIREN+pixel. We observe that \algname achieves better performance when an image contains less complex edges. We shows four examples in Figure \ref{exp_set14_eg}. The images of Face and Flower in Set14 contains more edges than the images of Pepper and PPT3. This implies that \algname~ reconstructs an image well when it contains more clear edges.   

\subsubsection{DIV2K}

DIV2K contains high resolution realistic images and we trained three models for validation dataset of the high resolution images of DIV2K which constains 100 samples. The image has higher resolution than $256\times 256$ and training bigger size of an image requires larger model and more computational time. Therefore, we down-sample the image to $256\times 256$ size. The result in Table \ref{table_psnr_ssim} shows that \algname~ has lower performance than SIREN+pixel. When we compare the performance of each samples, there is no case \algname~ outperforms SIREN+pixel. We hypothesize that this is due to the fact that high resolution images contains more complex edge patterns than original $256\times 256$ image and down-sampling does not contain clear edge information. 

\subsubsection{MNIST}

\begin{figure}[ht]
\includegraphics[width=8cm, trim = 0cm 0cm 0cm 1cm, clip]{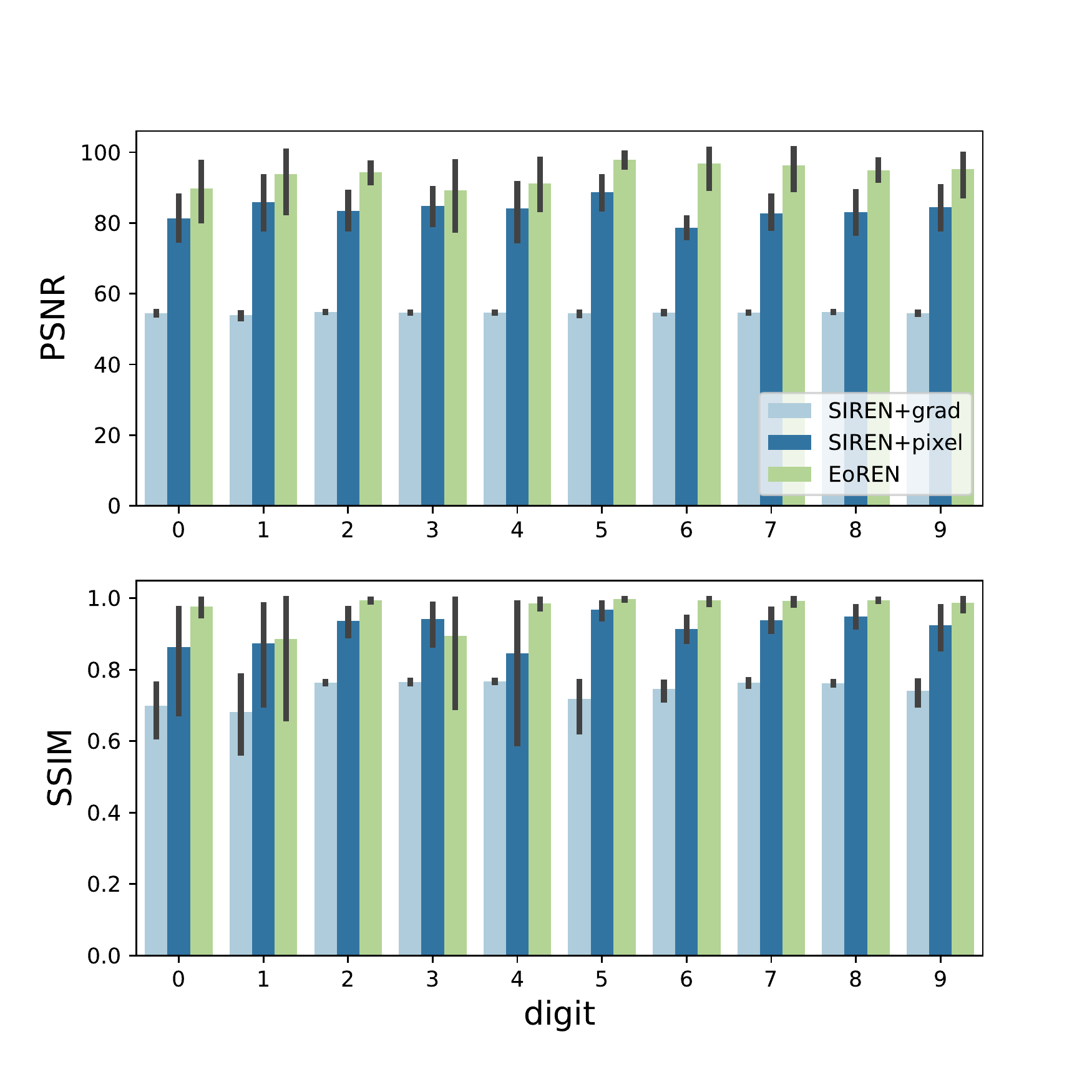}
\centering
\caption{The performance of \algname, SIREN with pixel fitting, and SIREN with gradient fitting for each digits. \algname~ outperforms other models for all digits. We test for 10 samples each. }
\label{exp_mnist}
\end{figure}

\begin{figure}[ht]
\includegraphics[width=8cm, trim = 0cm 0cm 0cm 0cm, clip]{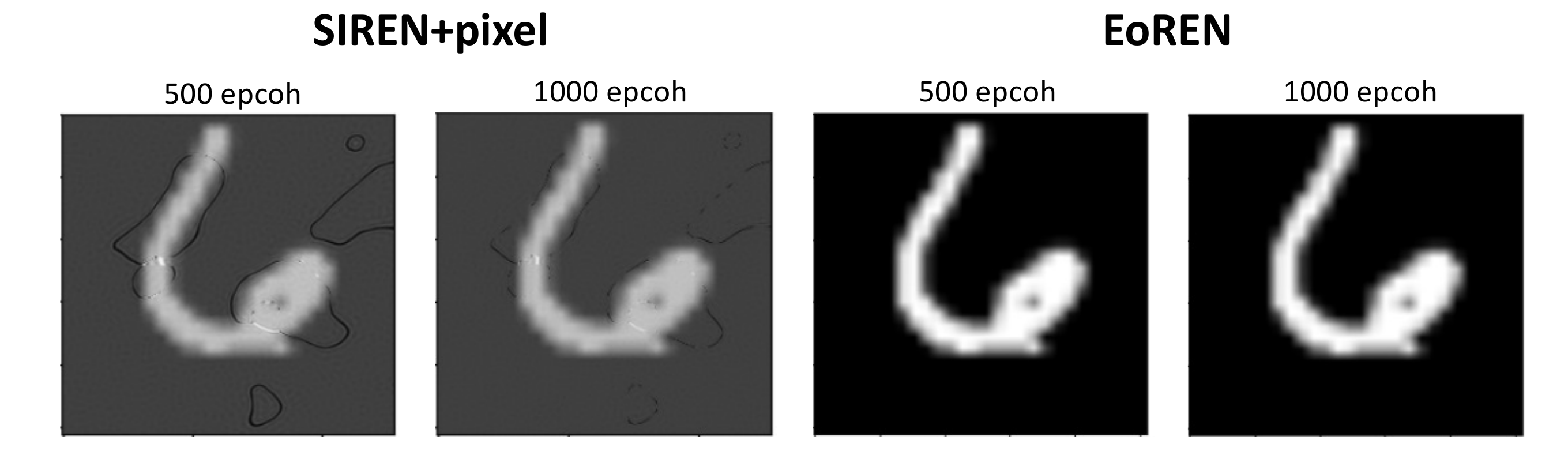}
\centering
\caption{Reconstruction sample of SIREN with pixel fitting and \algname~ for digit six for 500 and 1000 epochs. }
\label{exp_6_digit}
\end{figure}

Finally, we tested three models with MNIST dataset. As the results of Table \ref{table_psnr_ssim} and the stacked box plot in Figure \ref{exp_mnist} show, \algname~ significantly outperforms SIREN+pixel and SIREN+grad. We show an reconstruction example of digit six in Figure \ref{exp_6_digit}. When training SIREN with pixel fitting, there is a watermark on the reconstructed image, whereas, the reconstructed image of \algname has no watermark. When the \algname is trained without the channel-tuning module, it is equivalent to the SIREN+grad. As the performance of SIREN+grad suggests, the channel tuning module is necessary for the gradient fitting of image.

\subsection{Poisson Image Editing}

Two images $f_1$ and $f_2$ can be composited by the interpolation of them. The composite image $f_{1,2}$ can be trained using Possion image editing method \cite{perez2003poisson} which focuses on the edge of two images.  Poisson image editing technique uses addition of the gradients  of images and trained with single parametric function by 
\begin{gather}
    \mathcal{L}_{\mathrm{grad.}} = \int_{\Omega} || \nabla_\mathbf{x} \Phi(\mathbf{x}) - \nabla_\mathbf{x} f_{1,2}|| d\mathbf{x} \\ 
    f_{1,2} = \lambda f_1 (\mathbf{x}) + (1-\lambda) f_2 (\mathbf{x})  
    \label{equation:possion}
\end{gather}

When model $\Phi$ is fully trained by Equation \ref{equation:possion}, we get two edge based composited image. The proportion of two images is controlled by the constant $\lambda \in [0,1]$. In the case of \algname, there are already two hyperparameters $\gamma_1$ and $\gamma_2$ in a single image to control the training, the learning rate of the edge-oriented module and the offset-tuning module. Then, the composition of two images can be trained by two losses

\begin{gather}
    \mathcal{L}_{1,2}^{grad} = \lambda  \gamma_1  \mathcal{L}_{\mathrm{grad.}}^1   + (1-\lambda)  \gamma_1  \mathcal{L}_{\mathrm{grad.}}^2 \\
    \tilde{\mathcal{L}}_{1,2} =  \lambda  \gamma_2 \tilde{\mathcal{L}}^1  + (1-\lambda) \gamma_2 \tilde{\mathcal{L}}^2
\end{gather}

where $\mathcal{L}^i_{grad}$ is the loss of edge-oriented module of $f_i$ and $\tilde{\mathcal{L}}^i$ is the loss of offset-tuning module of $f_i$.
Therefore, there are total 3 number of independent hyperparameters to train composition. We show that by setting proper parameter, we can make the composition image with desired magnitude of edge and color. We test this Poisson image editing method for \algname.

\begin{figure}[ht]
\includegraphics[width=8cm, trim = 0cm 0cm 0cm 0cm, clip]{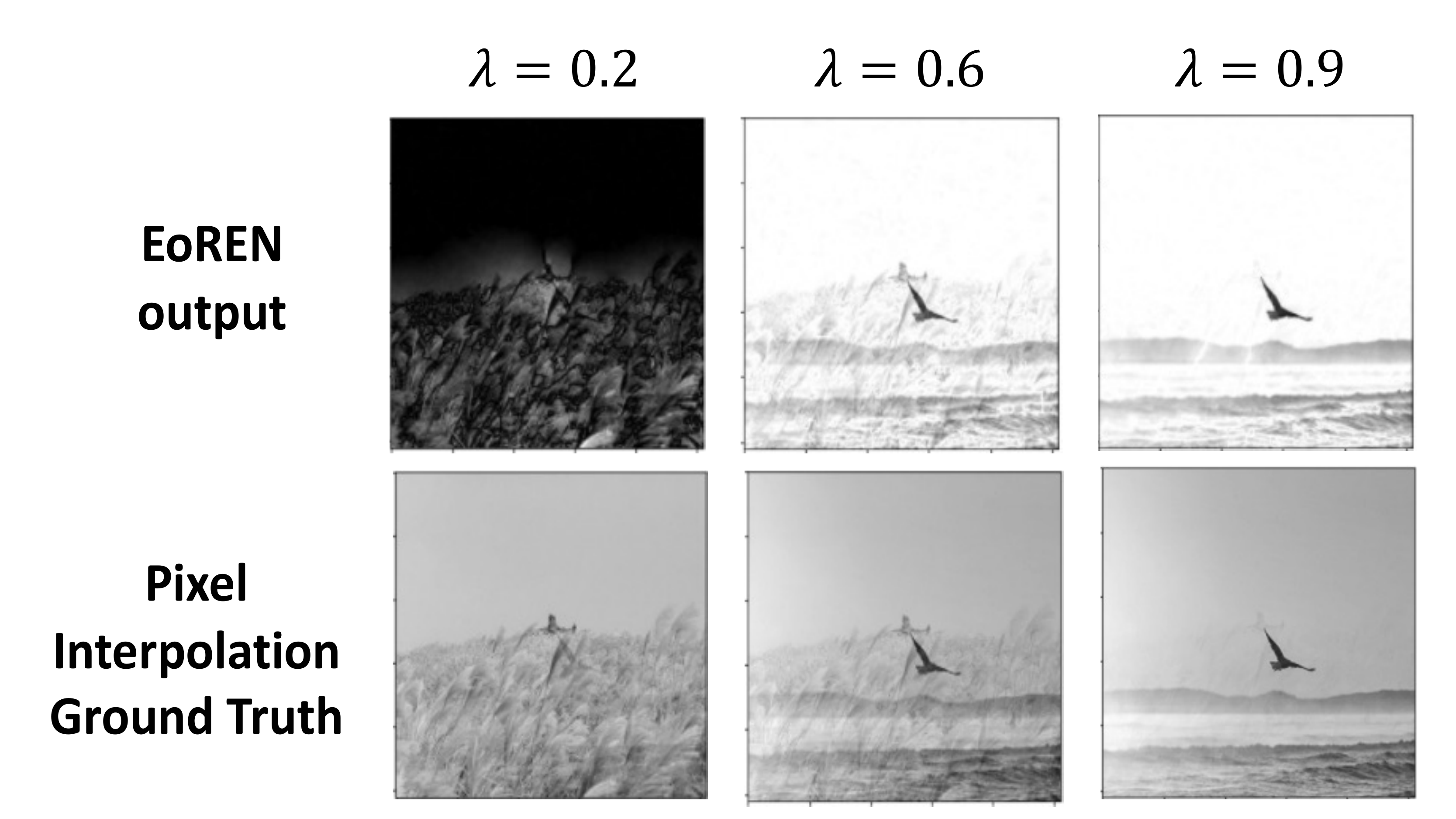}
\centering
\caption{The result of image composition for $\lambda=0.2, 0.6, 0.9$. \algname~ reconstruct the composited image successfully when  $\lambda$ is large indicating that \algname~ prefers an image with simple patterns.}
\label{exp_composition}
\end{figure}

We show the result of the image composition in Figure \ref{exp_composition}. Two images have different degree of complexity.The first one is an image of a field with reeds, and the second one is an image of an eagle flying against the sky and sea. The first image contains more edges than the second image and therefore more difficult to train with gradients. The result shows that \algname~ successfully combines two images when the portion of the second image is more than the first image. This implies that the channel tuning depends on the complexity of the edges.  

\section{Conclusions}

In this paper, we propose the Gradient Magnitude Adjustment algorithm to extract image gradients and the Edge-oriented Representation Network which disentangles the edge features and the channel-wise value features. We demonstrate that the implicitly represented image with the gradients adjusted by the GMA matches the original image well. In addition, we empirically show that the SIREN model fitted with pixel value makes a smooth image while the SIREN model fitted with gradients captures the edge parts well. Our proposed EoREN model is separated into two parts which makes both edges and values of the image be reconstructed well. \\
\indent Even though our model improves image reconstruction performance trained with implicit function, there still exist some challenges to be solved. The channel-tuning module highly depends on the performance of the edge-oriented module. Also, the model output is desaturated compared to the original image even though the distribution of pixel values have no significant difference. We expect significant improvements on the reconstruction performance by solving such problems.

{\bibliographystyle{ieee_fullname}
\bibliography{main}
}

\end{document}